\documentclass[runningheads]{llncs}

\usepackage{eccv}

\usepackage{eccvabbrv}

\usepackage{graphicx}
\usepackage{booktabs}
\usepackage{amsmath}
\usepackage{amssymb}
\usepackage{booktabs}
\usepackage{siunitx}
\usepackage{cite}
\usepackage{hhline}
\usepackage{subcaption}
\usepackage{csquotes}
\usepackage{multirow}
\usepackage{pifont}
\usepackage[super]{nth}
\usepackage{mathtools,bbm}
\usepackage{bm}
\usepackage{textcomp}
\usepackage{CJKutf8}
\usepackage{wrapfig}
\usepackage{float}
\usepackage{paralist}

\definecolor{darkergreen}{RGB}{21, 152, 56}
\definecolor{red2}{RGB}{252, 54, 65}

\newcolumntype{L}[1]{>{\raggedright\let\newline\\\arraybackslash\hspace{0pt}}m{#1}}
\newcolumntype{C}[1]{>{\centering\let\newline\\\arraybackslash\hspace{0pt}}m{#1}}
\newcolumntype{R}[1]{>{\raggedleft\let\newline\\\arraybackslash\hspace{0pt}}m{#1}}

\newcommand{\chinese}[1]{\begin{CJK}{UTF8}{bsmi}#1\end{CJK}}

\usepackage[accsupp]{axessibility}  

\usepackage{hyperref}

\usepackage{orcidlink}

\begin{document}

\title{Expressive Whole-Body 3D Gaussian Avatar} 


\author{Gyeongsik Moon\inst{1,2}\orcidlink{0000-0003-0610-7936} \and
Takaaki Shiratori\inst{2}\orcidlink{0000-0002-1012-415X} \and
Shunsuke Saito\inst{2}\orcidlink{0000-0003-2053-3472}}

\authorrunning{Moon et al.}

\institute{$^1$DGIST \hspace{26mm} $^2$Codec Avatars Lab, Meta \\
\email{mks0601@dgist.ac.kr} \hspace{4mm} \email{\{tshiratori,shunsukesaito\}@meta.com} \\
\url{https://mks0601.github.io/ExAvatar}
}

\maketitle

\begin{abstract}
Facial expression and hand motions are necessary to express our emotions and interact with the world.
Nevertheless, most of the 3D human avatars modeled from a casually captured video only support body motions without facial expressions and hand motions.
In this work, we present \textbf{ExAvatar}, an expressive whole-body 3D human avatar learned from a short monocular video.
We design ExAvatar as a combination of the whole-body parametric mesh model (SMPL-X) and 3D Gaussian Splatting (3DGS).
The main challenges are 1) a limited diversity of facial expressions and poses in the video and 2) the absence of 3D observations, such as 3D scans and RGBD images.
The limited diversity in the video makes animations with novel facial expressions and poses non-trivial.
In addition, the absence of 3D observations could cause significant ambiguity in human parts that are not observed in the video, which can result in noticeable artifacts under novel motions.
To address them, we introduce our hybrid representation of the mesh and 3D Gaussians.
Our hybrid representation treats each 3D Gaussian as a vertex on the surface with pre-defined connectivity information (\emph{i.e.}, triangle faces) between them following the mesh topology of SMPL-X.
It makes our ExAvatar animatable with novel facial expressions by driven by the facial expression space of SMPL-X.
In addition, by using connectivity-based regularizers, we significantly reduce artifacts in novel facial expressions and poses.
\end{abstract}

\section{Introduction}

Humans use all facial expressions, body motions, and hand motions to express our emotions and intentions, and interact with other people and objects.
In particular, facial expressions and hand gestures are one of the most powerful channels for non-verbal communication, and hand motions are necessary to interact with diverse types of objects.
Modeling the facial expression, body motion, and hand motion altogether is extremely challenging.
Several whole-body 3D human geometry models have been introduced~\cite{joo2018total,pavlakos2019expressive,xu2020ghum,alldieck2021imghum}.
Among them, SMPL-X~\cite{pavlakos2019expressive} is the most widely used one, which motivated a number of 3D whole-body pose estimation methods~\cite{choutas2020monocular,rong2021frankmocap,feng2021collaborative,moon2022accurate,zhang2023pymaf,li2023hybrikx,lin2023one,cai2023smpler} and benchmarks~\cite{patel2021agora}.

To represent 3D humans beyond the minimally clothed parametric models, personalized 3D human avatars have been recently studied.
The 3D human avatar is a representation that combines 3D geometry and the appearance of a certain person, which can be animated and rendered with novel poses.
However, most of existing 3D human avatars~\cite{peng2021neural,peng2021animatable,kwon2021neural,choi2022mononhr,chen2021animatable,jiang2022neuman,guo2023vid2avatar,jiang2023instantavatar,kocabas2023hugs,hu2023gaussianavatar} modeled from a casually captured video only support body motions without facial expressions and hand motions.
Their avatars bake facial expressions and hand poses, and animating them is not possible.
A recent work~\cite{shen2023x} introduced a whole-body avatar that supports animation with facial expressions, and body and hand poses; however, it requires 3D observations, such as 3D scans or RGBD images with highly accurate SMPL-X registrations, with diverse poses and facial expressions.
Such an assumption does not hold for the majority of casually captured videos in daily life.

\begin{figure}[t]
\begin{center}
\includegraphics[width=\linewidth]{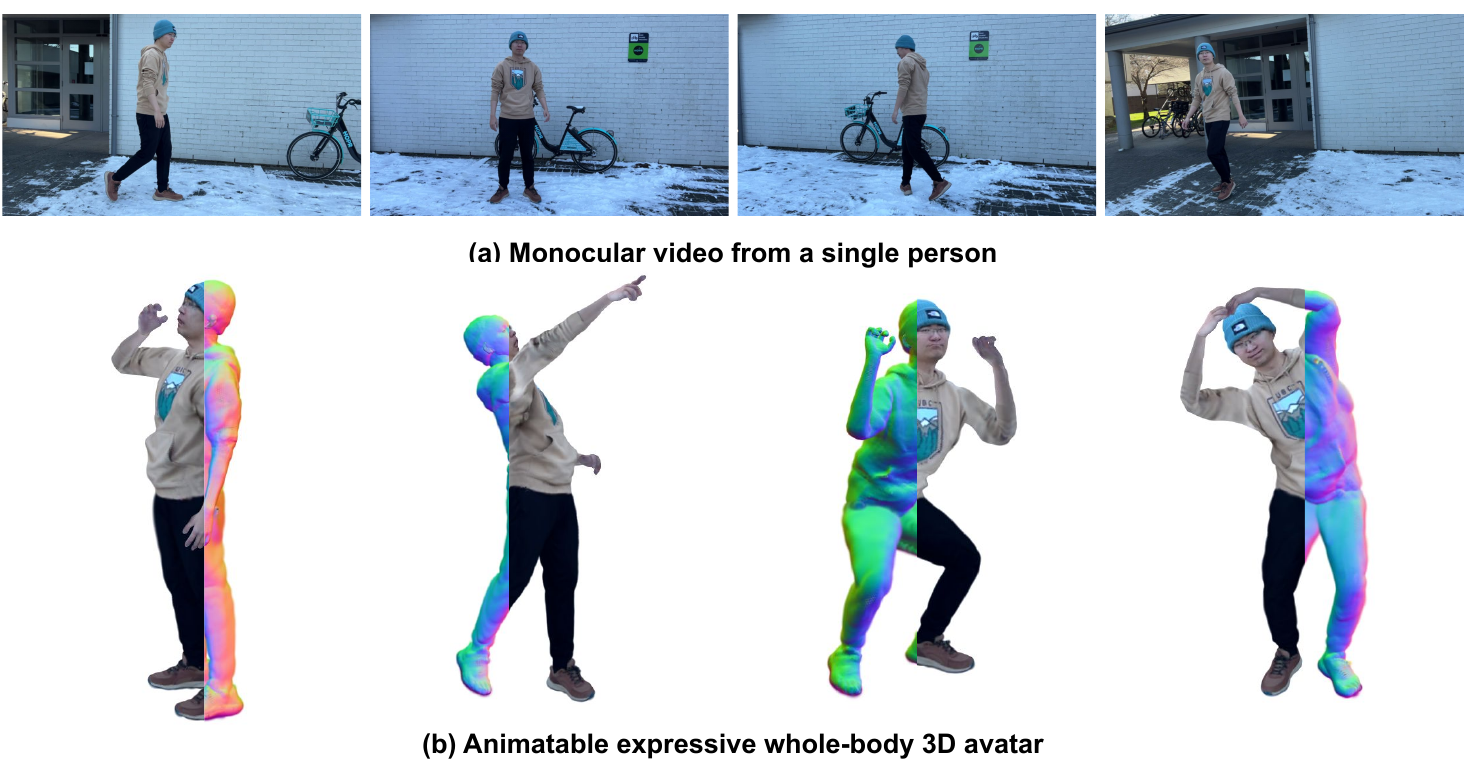}
\end{center}
\vspace*{-6mm}
\caption{
From (a) a monocular video from a single person, we create our (b) \textbf{ExAvatar}, an expressive whole-body 3D avatar, animatable with novel facial expression code, hand poses, and body poses of SMPL-X.
}
\vspace*{-5mm}
\label{fig:intro_teaser}
\end{figure}

We present \textbf{ExAvatar}, an expressive whole-body 3D human avatar that can be made from a short monocular video.
ExAvatar is designed as a combination of the whole-body 3D parametric model (SMPL-X)~\cite{pavlakos2019expressive} and 3D Gaussian Splatting (3DGS)~\cite{kerbl20233d}.
It utilizes the whole-body drivability of SMPL-X and the photorealistic and efficient rendering capability of 3DGS.
After the training, it is animatable with novel facial expression code and 3D pose of SMPL-X, as shown in Fig.~\ref{fig:intro_teaser}.
Despite its desired properties, modeling ExAvatar is an non-trivial task with the following two challenges: 1) a limited diversity of facial expressions and poses in the video and 2) the absence of 3D observations, such as 3D scans and RGBD videos.
The limited diversity in the video makes a drivability with novel facial expressions and poses non-trivial.
In addition, the absence of 3D observations creates ambiguity in the occluded human parts, exhibiting noticeable artifacts in novel facial expressions and poses.

To address them, we propose a novel hybrid representation of the surface mesh and 3D Gaussians in ExAvatar.
Our hybrid representation treats each 3D Gaussian as a vertex on the surface, where the vertices have pre-defined connectivity  (\emph{i.e.}, triangle faces) between them following the mesh topology of SMPL-X.
Existing volumetric avatars~\cite{peng2021neural,peng2021animatable,kwon2021neural,choi2022mononhr,chen2021animatable,jiang2022neuman,guo2023vid2avatar,jiang2023instantavatar,shen2023x} do not have the connectivity by the definition.
Also, previous 3DGS-based~\cite{kocabas2023hugs,hu2023gaussianavatar} works consider a set of 3D Gaussian points as a point cloud without considering the connectivity between them.

Using our hybrid representation, our ExAvatar becomes fully compatible with the facial expression space of SMPL-X.
Therefore, it can be driven with any facial expression code of SMPL-X \emph{even from a short monocular video without diverse facial expressions}.
As our 3D Gaussians share the exactly same mesh topology with SMPL-X, we simply add the vertex offsets to our 3D Gaussian points to move them according to the facial expression as in FLAME~\cite{li2017learning} and SMPL-X~\cite{pavlakos2019expressive}.
Hence, unlike previous works~\cite{shen2023x}, our drivability of the facial expression is not strictly limited by the number of training frames (\emph{e.g.}, 30 seconds of a short video).

Another benefit is that we can significantly reduce artifacts in novel facial expressions and poses using connectivity-based regularizers.
As the pose diversity in the training set is very limited, there could be human parts that are not observed at all in the video.
Without 3D observations, the ambiguous human parts could suffer from artifacts in novel poses.
While several point-based regularizers (e.g., L2 regularization of the underlying SMPL/SMPL-X template mesh) have been proposed~\cite{hu2023gaussianavatar}, they do not consider \emph{connectivity} between vertices.
Such a lack of connectivity could introduce floating 3D Gaussians.
By considering the connectivity, we can naturally enforce local similarity, significantly reducing artifacts.

Throughout our experiments, 
our method substantially outperforms all previous 3D human avatars in various benchmarks.
Our contributions can be summarized as follows.
\begin{itemize}
\item We present ExAvatar, an expressive whole-body 3D human avatar that can be made from a short monocular video without requiring 3D observations.
\item We propose a hybrid representation of the surface mesh and 3D Gaussians.
It allows ExAvatar to be animated with any novel facial expression code of SMPL-X even from a short monocular video without diverse facial expressions.
\item Using connectivity information between 3D Gaussians, we significantly reduce artifacts, especially in novel facial expressions and poses.
\end{itemize}

\section{Related works}

\noindent\textbf{3D human avatars.}
Various 3D representations are used for modeling 3D human avatars.
Alldieck~\etal~\cite{alldieck2018video} extended SMPL mesh with per-vertex offsets.
Bagautdinov~\etal~\cite{bagautdinov2021driving} achieved high-fidelity results using conditional variational autoencoder, which can be animated with incomplete driving signals.
Remelli~\etal~\cite{remelli2022drivable} propose to use texel-aligned features, a localized representation.
Motivated by neural radiance fields~\cite{mildenhall2021nerf}, many volumetric and implicit representation-based avatars have been introduced.
Peng~\etal~\cite{peng2021neural,peng2021animatable} created a 3D human avatar from a capture studio, which provides accurate 3D pose and multi-view images for supervision.
Kwon~\etal~\cite{kwon2021neural} improved the previous works by utilizing vertex-aligned features.
Shen~\etal~\cite{shen2023x} created a whole-body 3D avatar, which supports whole-body animation with facial expression, from their capture studio dataset.
In contrast to the above works that make 3D avatars from capture studios, recent works focus on making 3D avatars from a short monocular video without requiring 3D observations, such as 3D scans, RGBD images, or multi-view images.
Jiang~\etal~\cite{jiang2022neuman} introduced a dataset and method for making a 3D human avatar from a short monocular video taken from in-the-wild environments.
Guo~\etal~\cite{guo2023vid2avatar} proposed a system that can decompose a scene and human with self-supervised learning.
Jiang~\etal~\cite{jiang2023instantavatar} introduced a system that can make a 3D human within several minutes.
Recently introduced 3DGS~\cite{kerbl20233d}, which achieves both powerful and efficient rendering capability, motivated several 3DGS-based avatars~\cite{kocabas2023hugs,hu2023gaussianavatar,liu2024gea}.
Kocabas~\etal~\cite{kocabas2023hugs} use the triplane for creating 3D avatar.
Hu~\etal~\cite{hu2023gaussianavatar} introduced a robust system that takes a positional map of a posed SMPL mesh.
Moon~\etal~\cite{Moon_2024_CVPR_UHM} presented universal hand model (UHM) to create authentic hand avatars from a phone scan.
Chen~\etal~\cite{chen2024urhand} extended UHM of Moon~\etal~\cite{Moon_2024_CVPR_UHM} for the relightability.

Except for a few works~\cite{bagautdinov2021driving,shen2023x,liu2024gea}, most of the above works only support body motions without hand motions and facial expressions.
X-Avatar~\cite{shen2023x} supports whole-body animation including facial expressions; however, it has two limitations.
First, it requires diverse facial expressions with accurate 3D geometry registration in videos to create avatars.
This is because they cannot directly utilize the facial expression space of FLAME~\cite{li2017learning}/SMPL-X~\cite{pavlakos2019expressive}.
They need to transform the mesh-based facial expression space of FLAME~\cite{li2017learning}/SMPL-X~\cite{pavlakos2019expressive} to their implicit representation using learnable modules.
To train the transformation module, they need training data with sufficiently diverse facial expressions and accurate 3D geometry registrations.
Second, it requires 3D observations, such as 3D scans or RGBD images with accurate SMPL-X registrations, for the training, hard to obtain from in-the-wild environments.
Due to the above two reasons, X-Avatar~\cite{shen2023x} is hard to apply to practical settings, such as short monocular videos.
Liu~\etal~\cite{liu2024gea} proposed another whole-body 3D avatar; however, their avatar is not animated with novel facial expressions.

\noindent\textbf{Whole-body 3D human modeling and perception.}
Modeling face, body, and hands at the same time is an extremely challenging problem as each human part has its own different characteristics.
Several whole-body 3D human models have been introduced~\cite{joo2018total,pavlakos2019expressive,xu2020ghum,alldieck2021imghum}, which model 3D geometry of minimally clothed humans.
They are parametric models, parameterized by 3D poses, facial expression code, and shape parameter.
Among them, SMPL-X~\cite{pavlakos2019expressive} is the most widely used one due to its completeness.
Motivated by the optimization baseline~\cite{pavlakos2019expressive} and benchmarks~\cite{patel2021agora}, a number of 3D whole-body pose estimation methods~\cite{choutas2020monocular,rong2021frankmocap,feng2021collaborative,moon2022accurate,zhang2023pymaf,li2023hybrikx,lin2023one,cai2023smpler} have been introduced.
\begin{figure}[t]
\begin{center}
\includegraphics[width=\linewidth]{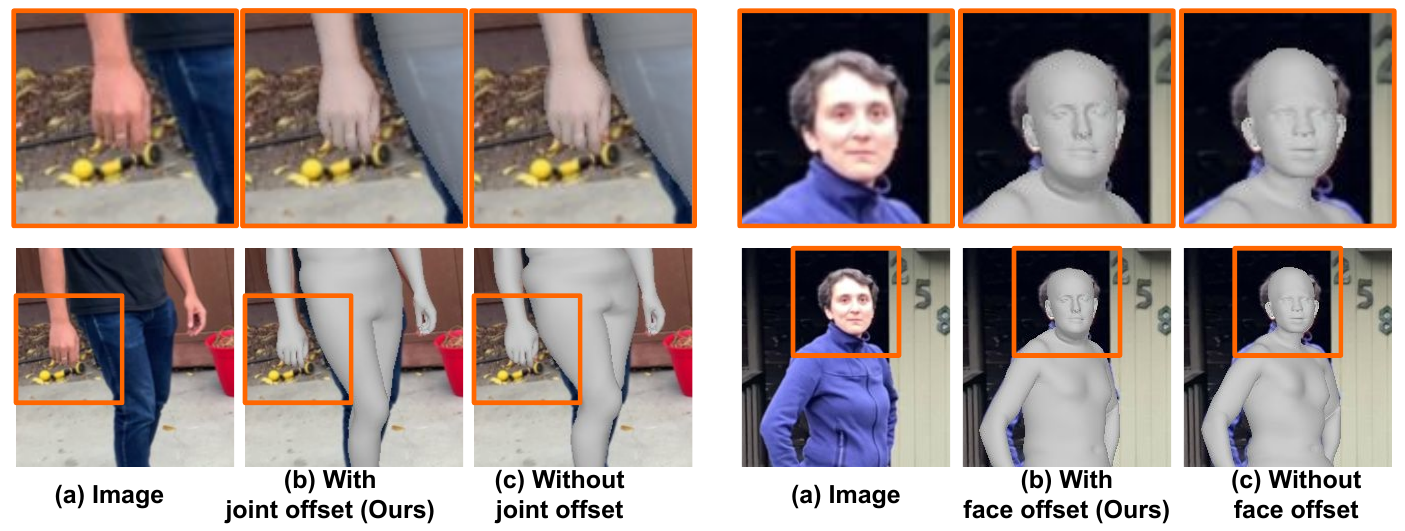}
\end{center}
\vspace*{-6mm}
\caption{
The effectiveness of our joint offset $\Delta \mathbf{J}$ and face offset $\Delta \mathbf{V}_\text{face}$.
They are necessary for the accurate registration of hands and face, which results in accurate co-registration of the whole body.
}
\vspace*{-5mm}
\label{fig:joint_face_offsets}
\end{figure}

\section{ExAvatar}

\subsection{Accurate co-registration of SMPL-X}~\label{subsec:preprocessing}
We assume videos, which usually consist of 30 seconds of frames, are taken from an in-the-wild environment.
The video is from a single person with a natural backgrounds.
Before training our ExAvatar, we first preprocess the video.
Following previous works~\cite{jiang2022neuman,guo2023vid2avatar}, we first run an off-the-shelf SMPL-X regressor~\cite{li2023hybrikx} and 2D pose estimator~\cite{mmpose2020} to all frames.
Then, we additionally fit the regressed SMPL-X parameters (\emph{i.e.}, 3D poses $\theta \in \mathbb{R}^{55 \times 3}$, shape parameter $\beta \in \mathbb{R}^{100}$, and facial expression code $\psi \in \mathbb{R}^{50}$) and 3D translation $\mathbf{t}$ to the estimated 2D pose of each frame.
The shape parameter is shared across all frames as all frames are from the same person.

\begin{wrapfigure}{r}{0.45\linewidth}[t]
\vspace{-3.5em}
\begin{center}
\includegraphics[width=\linewidth]{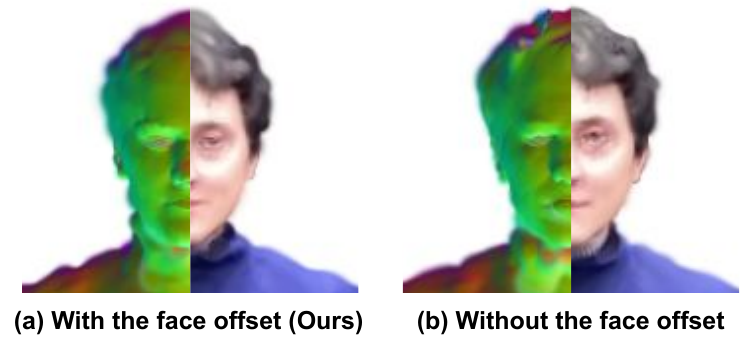}
\end{center}
\vspace{-4mm}
\caption{
Without the face offset $\Delta \mathbf{V}_\text{face}$, the final 3D geometry of the avatar becomes totally inauthentic and inaccurate.
For each setting, normals of 3D Gaussian points and colors are used for the rendering.
}
\label{fig:face_offset_effect}
\end{wrapfigure}
One challenge during registering SMPL-X parameters to a video is accurate \emph{co-registration} of body, hands, and face, unique challenges of the whole-body avatar.
The registration of hands and face can be negatively affected by a limited expressiveness of SMPL-X and registration accuracy of the body, which can limit the co-registration accuracy.
To achieve the \emph{accurate co-registration} of body, hands, and face, we introduce two optimizable offsets initialized with zero and shared across all frames.
Both offsets are identity (ID)-dependent offsets and are not dependent on the poses and facial expressions.
Hence, they are added to the T-pose template mesh of SMPL-X before performing the linear blend skinning (LBS).

First, we introduce joint offset $\Delta \mathbf{J}$, added to the joints in the T-pose space of SMPL-X.
The joint offset $\Delta \mathbf{J}$, which affect both 3D skeleton and surface, are especially helpful to fit hands more perfectly as the shape parameter of SMPL-X has limited coverage of 3D hand skeleton, as shown in Fig.~\ref{fig:joint_face_offsets} left.
Second, we introduce face offset $\Delta \mathbf{V}_\text{face}$, per-vertex offset of the face region of SMPL-X, added to the face vertex in the T-pose space of SMPL-X.
To optimize the face offset $\Delta \mathbf{V}_\text{face}$, we first fit the 3D face-only model (\emph{i.e.}, FLAME~\cite{li2017learning}) to 2D poses and images of the face by running DECA~\cite{feng2021learning} and further optimizing it to 2D poses.
Then, we optimize the face offset by making a summation of the face offset and 3D face vertices of SMPL-X close to fitted FLAME vertices.
The optimization is straightforward as the face region of SMPL-X has exactly the same topology as that of FLAME.
The rationale is that 1) the face-only model has higher expressiveness in its shape space than the whole-body model and 2) the registration of the face-only model is not affected by the body registration.
Fig.~\ref{fig:joint_face_offsets} right and ~\ref{fig:face_offset_effect} show the effectiveness of our face offsets.
Such a special treatment in the registration stage is greatly helpful for the final 3D avatar, not considered in previous whole-body avatars~\cite{shen2023x,liu2024gea}.
Please refer to the supplementary material about the details of the fitting.

\subsection{Architecture}~\label{subsec:architecture}
\begin{figure}[t]
\begin{center}
\includegraphics[width=\linewidth]{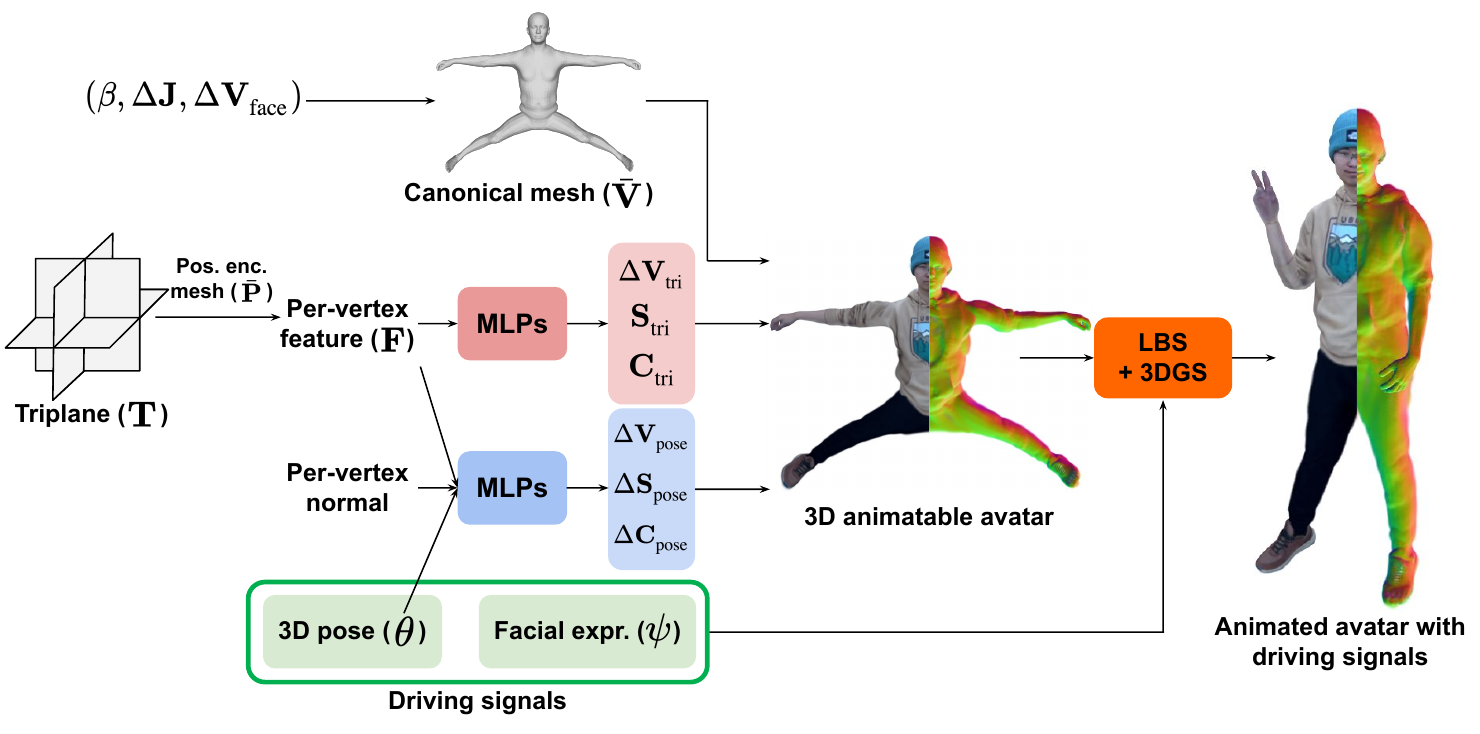}
\end{center}
\vspace*{-6mm}
\caption{
The architecture of our ExAvatar.
From the canonical mesh $\bar{\mathbf{V}}$, triplane $\mathbf{T}$, per-vertex normal, and 3D pose $\theta$, we build a 3D animatable avatar.
Then, with driving signals, 3D pose $\theta$ and facial expression code $\psi$ of SMPL-X~\cite{pavlakos2019expressive}, we animate the avatar and render it to the screen space with 3DGS~\cite{kerbl20233d}.
For the normal rendering, we calculate the normal vectors using the positions of 3D Gaussian points and mesh topology of SMPL-X.
}
\vspace*{-5mm}
\label{fig:architecture}
\end{figure}

Fig.~\ref{fig:architecture} shows the architecture of ExAvatar.
We model ExAvatar on top of a canonical 3D human mesh, denoted by $\bar{\mathbf{V}} \in \mathbb{R}^{N \times 3}$, where it has $N=167\text{K}$ upsampled vertices and 335K upsampled triangle faces.
To obtain it, we first pass the optimized SMPL-X shape parameter $\beta$, joint offsets $\Delta \mathbf{J}$, and face offsets $\Delta \mathbf{V}_\text{face}$ from Sec.~\ref{subsec:preprocessing}, and a pre-defined neural pose (\emph{i.e.}, \chinese{大}-pose) to the SMPL-X layer.
Then, we upsample it with the subdivision function of PyTorch3D~\cite{ravi2020accelerating}, which can upsample other 3D assets, such as facial expression blend shapes, in a consistent way.

\noindent\textbf{Per-vertex Gaussian assets regression.}
We initialize a learnable triplane~\cite{chan2022efficient} $\mathbf{T} \in \mathbb{R}^{3 \times C \times H \times W}$ with zero, where $C=32$, $H=128$, and $W=128$ represent channel dimension, height, and width of the triplane, respectively.
Then, we prepare a positional encoding mesh $\bar{\mathbf{P}} \in \mathbb{R}^{N \times 3}$ with a pre-defined neutral pose (\emph{i.e.}, \chinese{大}-pose) and zero shape parameter.
We upsample the positional encoding mesh with the above subdivision function, which produces the same mesh topology as the canonical mesh $\bar{\mathbf{V}}$.
We extract the per-vertex feature from the triplane by orthogonally projecting $\bar{\mathbf{P}}$ to each plane and performing the bilinear interpolation.
The triplane is useful as it naturally enforces similarity between close vertices.
In practice, we construct another triplane dedicated to the face, as the face requires detailed geometry and appearance modeling with a small physical size.
The reason for not using the canonical mesh $\bar{\mathbf{V}}$ for the feature extraction is that it keeps changing during the training as we further optimize the shape parameter $\beta$ and the joint offset $\Delta \mathbf{J}$ during the training.
If the position of a certain vertex changes, the extracted triplane feature of that vertex could be one that was from other vertices, which can make the training unstable.

The interpolated features from the triplane are concatenated, denoted by $\mathbf{F} \in \mathbb{R}^{N \times 96}$.
We pass $\mathbf{F}$ to two multi-layer perceptrons (MLPs), which regress 1) 3D offset $\Delta \mathbf{V}_\text{tri} \in \mathbb{R}^{N \times 3}$ and scale $\mathbf{S}_\text{tri} \in \mathbb{R}^{N \times 1}$ and 2) RGB values $\textbf{C}_\text{tri} \in \mathbb{R}^{N \times 3}$ for the 3DGS, respectively.
The MLPs are shared across all vertices.
Motivated by Hu~\etal~\cite{hu2023gaussianavatar}, for better generalization to novel viewpoints, we limit all Gaussian assets to be isotropic by limiting a degree of freedom of the scale to 1 and setting the rotation and opacity to identity and one, respectively.
Please refer to the supplementary material for the detailed architecture of the MLPs.
The regressed Gaussian assets (\emph{i.e.}, 3D offset, scale, and RGB values) are solely from the triplane, shared across all frames.
Hence, they represent identity (ID)-dependent and environment (\emph{e.g.}, lighting)-dependent assets without pose dependency as ID and environment are fixed in the input video, while pose changes for each frame.

To additionally model pose-dependent deformations, we employ two additional MLPs.
The first MLP takes $\mathbf{F}$ and 3D poses $\theta$ without the root pose and outputs 3D vertex offset $\Delta \mathbf{V}_\text{pose} \in \mathbb{R}^{N \times 3}$ and scale offset $\Delta \mathbf{S}_\text{pose} \in \mathbb{R}^{N \times 1}$.
The second MLP takes $\mathbf{F}$, 3D poses $\theta$ without the root pose, and the normal vector of each vertex and outputs RGB offset $\Delta \mathbf{C}_\text{pose} \in \mathbb{R}^{N \times 3}$.
The additional normal vector can 1) provide the view-dependent shading information to the network and 2) be useful to disentangle geometry and appearances~\cite{saito2019pifu,guo2023vid2avatar}.
Thanks to our hybrid representation, we can easily obtain the per-vertex normal vector by averaging normals of triangle faces that include the vertex.
Instead of directly predicting pose-dependent Gaussian assets, ours output pose-dependent offsets.
This is helpful for the generalization to novel poses as Gaussian assets solely from the triplane already have reasonable expressiveness, which makes the role of the pose-dependent Gaussian assets small.
Such a design is especially important when making 3D avatars from a short video like ours as limited pose diversity makes generalization to novel poses challenging.

\subsection{Animation and rendering}

\begin{figure}[t]
\begin{center}
\includegraphics[width=0.77\linewidth]{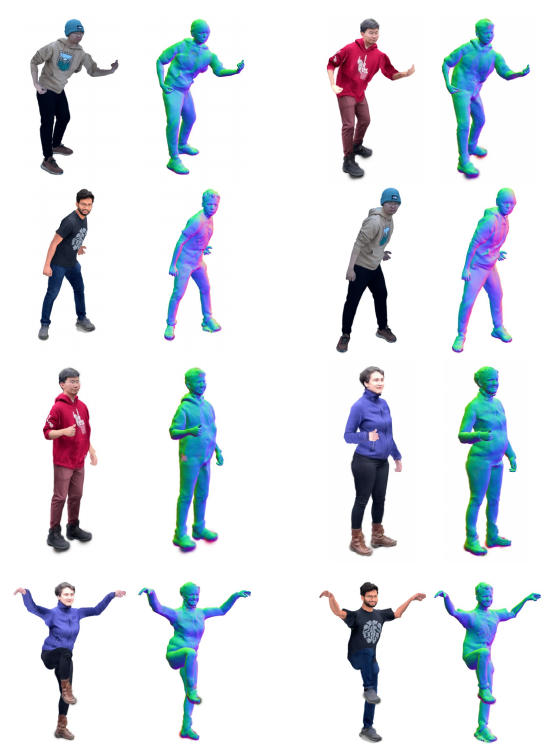}
\end{center}
\vspace*{-7mm}
\caption{
Our animated expressive whole-body avatars, made from monocular videos of NeuMan dataset~\cite{jiang2022neuman}.
Avatars of each row are animated with the same facial expression code $\psi$ and 3D pose $\theta$ of SMPL-X.
}
\vspace*{-5mm}
\label{fig:animation}
\end{figure}

Fig.~\ref{fig:animation} shows examples of our animated and rendered avatars, made from short monocular videos.

\noindent\textbf{Animation.}
We need to animate Gaussian points from the canonical space with given facial expression code $\psi$ and 3D poses $\theta$ of SMPL-X.
To this end, we first replace the pose-dependent vertex offset $\Delta \mathbf{V}_\text{pose}$ of hand and face vertices to those of SMPL-X.
This is because the hand and face are often naked; hence, we can directly utilize vertex offsets of SMPL-X.
Then, we add vertex offsets from facial expression code $\psi$ of SMPL-X  to face vertices.
\emph{By directly using the facial expression offsets of SMPL-X, we do not have to learn a new facial expression space.}
Such a direct utilizing is from our \emph{hybrid representation of the mesh and 3D Gaussians}.
The below equations describe the above deformations in the \emph{canonical space}.
\begin{equation}~\label{eq:animation_canonical_tri}
\bar{\mathbf{V}}_\text{tri} = \bar{\mathbf{V}} + \Delta \mathbf{V}_\text{tri} + \Delta \mathbf{V}_\text{expr},
\end{equation}
\begin{equation}~\label{eq:animation_canonical_pose}
\bar{\mathbf{V}}_\text{pose} = \bar{\mathbf{V}} + \Delta \mathbf{V}_\text{tri} + \Delta \mathbf{V}_\text{pose} + \Delta \mathbf{V}_\text{expr},
\end{equation}
where $\Delta \mathbf{V}_\text{expr}$ represents the facial expression offset of SMPL-X, obtained from the facial expression code $\psi$.
Then, for the body vertices, we take the skinning weight of the nearest vertices from downsampled $\bar{\mathbf{V}}$, while for the hand and face vertices, we use the original skinning weight of the vertices.
This is because, for the body vertices, their semantic meaning could change due to the cloth geometry.
The final animated geometry, $\mathbf{V}_\text{tri}$ and $\mathbf{V}_\text{pose}$, are represented with below equations.
\begin{equation}~\label{eq:animation_posed}
\mathbf{V}_\text{tri} = \text{LBS}(\bar{\mathbf{V}}_\text{tri}, \theta, \mathbf{W}_\text{tri}) \quad \text{and} \quad  \mathbf{V}_\text{pose} = \text{LBS}(\bar{\mathbf{V}}_\text{pose}, \theta, \mathbf{W}_\text{pose}),
\end{equation}
where $\mathbf{W}_\text{tri}$ and $\mathbf{W}_\text{pose}$ represent the skinning weight of $\bar{\mathbf{V}}_\text{tri}$ and $\bar{\mathbf{V}}_\text{pose}$, respectively.

\noindent\textbf{Rendering.}
To render animated 3D geometry, we use 3DGS rendering pipeline~\cite{kerbl20233d} like the below equations.
\begin{equation}~\label{eq:3dgs_render_tri}
\mathbf{I}_\text{tri} = f(\mathbf{V}_\text{tri}, \text{exp}(\mathbf{S}_\text{tri}), \mathbf{C}_\text{tri}, \mathbf{K}, \mathbf{E}),
\end{equation}
\begin{equation}~\label{eq:3dgs_render_pose}
\mathbf{I}_\text{pose} = f(\mathbf{V}_\text{pose}, \text{exp}(\mathbf{S}_\text{tri} + \Delta \mathbf{S}_\text{pose}), \mathbf{C}_\text{tri} + \Delta \mathbf{C}_\text{pose}, \mathbf{K}, \mathbf{E}),
\end{equation}
where $f$, $\mathbf{K}$, and $\mathbf{E}$ represent rendering function of 3DGS, camera intrinsic, and extrinsic matrices, respectively.
As described above, following Hu~\etal~\cite{hu2023gaussianavatar}, we restrict all Gaussian assets to isotropic for better generalization; hence, rotation and opacity of all Gaussian points are set to identity and one, respectively, not described in the equations.

\subsection{Loss functions}\label{subsec:loss_functions}
During the training of our ExAvatar, we optimize the triplane $\mathbf{T}$, MLPs for the regression of Gaussian assets in Sec.~\ref{subsec:architecture}, 3D pose $\theta$ of each frame, facial expression code $\psi$ of each frame, 3D translation $\mathbf{t}$ of each frame, the shape parameter $\beta$, and the joint offset $\Delta \mathbf{J}$.
Also, we simultaneously optimize a 3DGS for the background following the original implementation~\cite{kerbl20233d} by segmenting out human regions using human masks from an off-the-shelf human segmentation model~\cite{he2017mask}.
Modeling background simultaneously produces better foreground mask~\cite{guo2023vid2avatar}, as estimated masks often have errors, especially on hand parts.
We denote rendered images from a combination of Eq.~\ref{eq:3dgs_render_tri} and the 3DGS for the background by $\mathbf{I}_\text{tri}^*$.
Likewise, we denote rendered images from a combination of Eq.~\ref{eq:3dgs_render_pose} and the 3DGS for the background by $\mathbf{I}_\text{pose}^*$.
To train ExAvatar, we minimize the below loss functions.

\noindent\textbf{Image loss.}
Following 3DGS~\cite{kerbl20233d}, we minimize $L1$ distance, 1 - SSIM, and LPIPS~\cite{zhang2018unreasonable} between rendered images (\emph{i.e.}, $\mathbf{I}_\text{tri}^*$ and $\mathbf{I}_\text{pose}^*$) and the captured image. 
We found that the additional LPIPS is helpful for sharper textures.
To save the computation, we compute the image loss after cropping the human region.

\begin{figure}[t]
\begin{center}
\includegraphics[width=\linewidth]{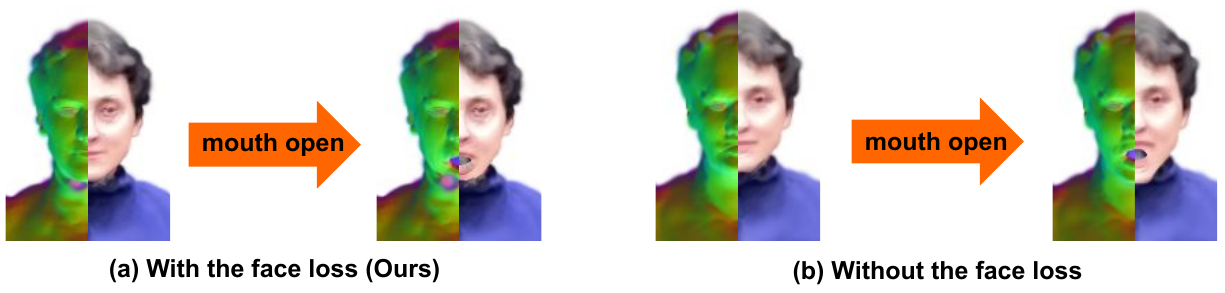}
\end{center}
\vspace*{-6mm}
\caption{
The effectiveness of our face loss.
Without the face loss, the geometry and texture of the face are not consistent, which makes significant artifacts when driving.
The right one (b) shows that without the face loss, when the mouth is opened, the upper lip remains at the same position, while only below lip is opened.
}
\vspace*{-5mm}
\label{fig:face_loss_effect}
\end{figure}

\begin{figure}[t]
\begin{center}
\includegraphics[width=\linewidth]{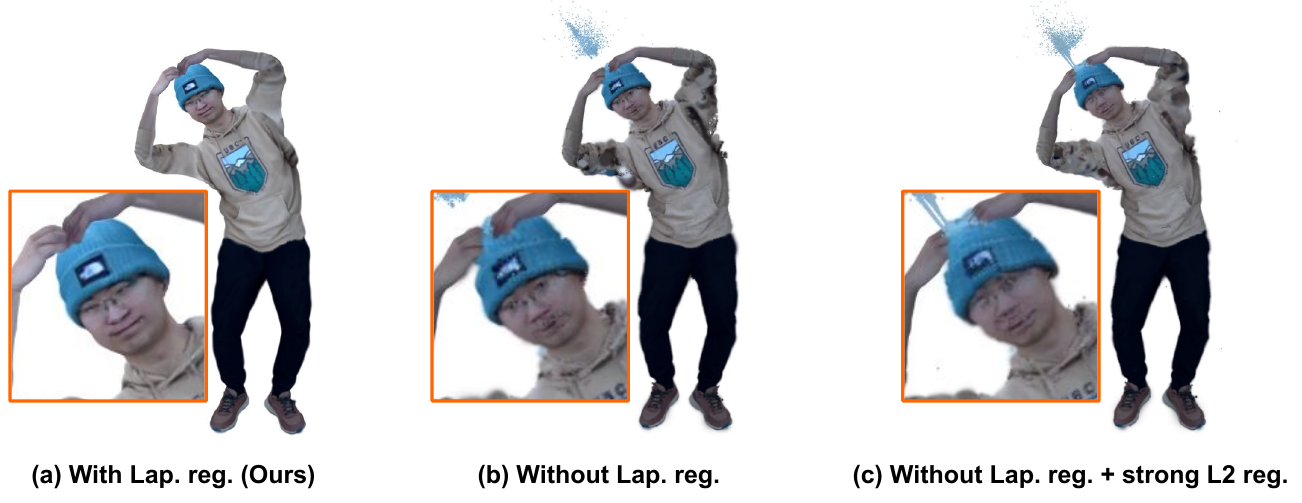}
\end{center}
\vspace*{-6mm}
\caption{
The effectiveness of the Laplacian regularizer, which makes 3D avatars in novel facial expressions and poses greatly stable.
On the other hand, the widely used $L2$ regularizer to the distance from SMPL-X surface to 3D Gaussian points suffers from severe artifacts.
We successfully incorporated the Laplacian regularizer using our hybrid representation of the surface mesh and 3D Gaussians.
}
\vspace*{-5mm}
\label{fig:lap_reg_effect}
\end{figure}

\noindent\textbf{Face loss.}
Unlike other human parts, the face has its unique characteristics as there should be a strong consistency between geometry and texture.
For example, lip geometry usually has reddish textures.
If other face geometry has lip textures, in novel facial expressions or jaw poses, the lip would not properly change, which can lead to significant artifacts, as shown in Fig.~\ref{fig:face_loss_effect} (b).
Simply minimizing the above image loss does not guarantee the consistency between the geometry and texture of the face region.
To enforce the consistency, we minimize the $L1$ distance between the rendered face image with a standard differentiable mesh renderer and the captured image, where the texture for the mesh renderer is prepared by averaging the unwrapped UV texture of the face-only model~\cite{li2017learning} registrations from Sec.~\ref{subsec:preprocessing}.
The UV texture is fixed, and the positions of 3D Gaussian points of the face region are adjusted to minimize the loss function.
Fig.~\ref{fig:face_loss_effect} (a) shows the effectiveness of our face loss functions.
Thanks to our hybrid representation of the mesh and 3D Gaussians, such a mesh-based loss function can be easily incorporated into our system.

\noindent\textbf{Regularizers.}
Due to the limited pose diversity in the training set, there can be human parts that are not observed in the video.
Such human parts suffer from occlusion ambiguity, which could result in artifacts in novel facial expressions and poses.
In addition, to utilize the facial expression offsets of SMPL-X, we need to make the face geometry similar to that of SMPL-X.
To address them, we utilize connectivity-based regularizers (\emph{i.e.}, Laplacian regularizer), motivated by the mesh modeling works~\cite{liu2019soft,moon2020deephandmesh}.
Fig.~\ref{fig:lap_reg_effect} shows that our connectivity-based regularizer significantly reduces artifacts in novel facial expressions and poses.
We minimize the difference of the 1) Laplacian of deformed 3D Gaussian points in the canonical space (\emph{i.e.}, $\bar{\mathbf{V}}_\text{tri}$ and $\bar{\mathbf{V}}_\text{pose}$) and 2) Laplacian of the canonical mesh $\bar{\mathbf{V}}$.
In this way, we can easily encourage the local similarity between 3D Gaussian points, which can prevent floating 3D Gaussians.
In particular, our connectivity-based regularizer is much more effective than the widely used $L2$ regularizer~\cite{hu2023gaussianavatar} that simply penalizes distance between 3D Gaussian points and underlying template mesh without considering the connectivity information. 
Due to our hybrid representation of the mesh and 3D Gaussians, the Laplacian regularizer, widely used in mesh modeling works, can be easily included in our system.
In addition to regularizing the 3D positions of 3D Gaussian points, we compute the same Laplacian regularizer for the scales and RGBs of 3D Gaussian points.
For other regularizers, please refer to the supplementary material.

\section{Experiments}

\subsection{Datasets}

\noindent\textbf{NeuMan.}
NeuMan~\cite{jiang2022neuman} provides several short monocular videos taken from in-the-wild environments.
Each video contains a single person walking around for about 15 seconds.
Following previous works~\cite{hu2023gaussianavatar} we use \emph{bike, citron, jogging, and seattle} videos that exhibit most human body regions and contain minimal blurry images.
We follow their official training and testing splits.

\noindent\textbf{X-Humans.}
X-Humans~\cite{shen2023x} provides 3D scans and RGBD videos of multiple subjects, captured from a studio.
Compared to NeuMan, X-Humans has more diverse facial expressions and hand poses.
There are two experimental protocols: 1) using 3D scans and 2) using RGBD images for creating avatars.
We create our avatar only with monocular RGB videos \emph{without depthmaps} and compare ours against previous works~\cite{shen2023x} that use RGBD videos.
We use \emph{0028, 0034, and 0087} subjects as their pre-trained weights of the RGBD protocol are publicly available.
We follow their official training and testing splits.

\begin{table}[t]
\setlength{\tabcolsep}{1pt}
\begin{minipage}{.45\linewidth}
\centering
\caption{Comparisons of 3D human avatars on the test set of NeuMan~\cite{jiang2022neuman} Rendered backgrounds are considered in the evaluation.
Only our ExAvatar supports face and hand animations.}
\label{table:compare_sota_neuman_w_bkg}
\vspace*{-3mm}
\scalebox{0.8}{
\begin{tabular}{C{3.0cm}|C{1.2cm}C{1.2cm}C{1.2cm}}
\specialrule{.1em}{.05em}{.05em}
Methods & PSNR\textuparrow & SSIM\textuparrow & LPIPS\textdownarrow \\ \hline
NeuMan~\cite{jiang2022neuman} & 24.22 & 0.77 & 0.27 \\  
Vid2Avatar~\cite{guo2023vid2avatar} & 15.41 & 0.53 & 0.66 \\ 
HUGS~\cite{kocabas2023hugs} & 25.17 & 0.83 & 0.16 \\
\textbf{ExAvatar (Ours)} & \textbf{27.47} & \textbf{0.90} & \textbf{0.10} \\
\specialrule{.1em}{.05em}{.05em}
\end{tabular}
}
\end{minipage}\hfill
\begin{minipage}{.50\linewidth}
\centering
\caption{Comparisons of 3D human avatars on the test set of NeuMan~\cite{jiang2022neuman}. Rendered backgrounds are \textbf{not} considered in the evaluation.
Only our ExAvatar supports face and hand animations.}
\label{table:compare_sota_neuman_wo_bkg}
\vspace*{-3mm}
\scalebox{0.8}{
\begin{tabular}{C{3.0cm}|C{1.2cm}C{1.2cm}C{1.2cm}}
\specialrule{.1em}{.05em}{.05em}
Methods & PSNR\textuparrow & SSIM\textuparrow & LPIPS\textdownarrow \\ \hline
HumanNeRF~\cite{weng2022humannerf} & 27.06 & 0.967 & 0.019 \\
InstantAvatar~\cite{jiang2023instantavatar} & 28.47 & 0.972 & 0.028 \\
NeuMan~\cite{jiang2022neuman} & 29.32 & 0.972 & 0.014 \\  
Vid2Avatar~\cite{guo2023vid2avatar} & 30.70 & 0.980 & 0.014 \\
GaussianAvatar~\cite{hu2023gaussianavatar} & 29.94 & 0.980 & 0.012 \\
3DGS-Avatar~\cite{qian20243dgs} & 28.99 & 0.974 & 0.016 \\
\textbf{ExAvatar (Ours)} & \textbf{34.80} & \textbf{0.984} & \textbf{0.009} \\
\specialrule{.1em}{.05em}{.05em}
\end{tabular}
}
\end{minipage}
\end{table}

\begin{figure}[t]
\begin{center}
\includegraphics[width=\linewidth]{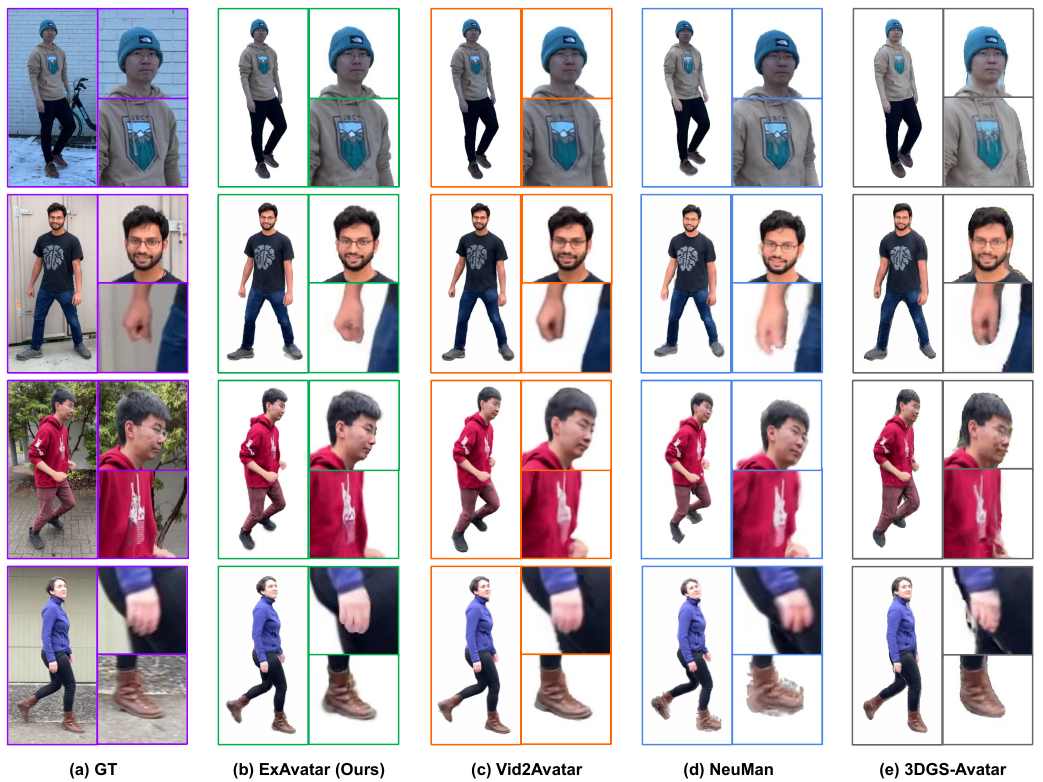}
\end{center}
\vspace*{-7mm}
\caption{
Qualitative comparison of our ExAvatar, Vid2Avatar~\cite{guo2023vid2avatar}, NeuMan~\cite{jiang2022neuman}, and 3DGS-Avatar~\cite{qian20243dgs} on the test set of Neuman~\cite{jiang2022neuman}.
}
\vspace*{-5mm}
\label{fig:compare_sota_neuman}
\end{figure}

\subsection{Comparison to state-of-the-art methods}
Tab.~\ref{table:compare_sota_neuman_w_bkg} and ~\ref{table:compare_sota_neuman_wo_bkg} show that our ExAvatar achieves the best results on NeuMan~\cite{jiang2022neuman} dataset regardless of whether the rendered background pixels are included or not.
All numbers are from papers~\cite{kocabas2023hugs,hu2023gaussianavatar} except NeuMan~\cite{jiang2022neuman}, Vid2Avatar~\cite{guo2023vid2avatar}, and 3DGS-Avatar~\cite{qian20243dgs} of Tab.~\ref{table:compare_sota_neuman_wo_bkg}, measured with their officially released code.
To exclude background pixels, we used an off-the-shelf segmentation network~\cite{he2017mask} following GaussianAvatar~\cite{hu2023gaussianavatar}.
Following previous works~\cite{chen2021animatable,hu2023gaussianavatar,qian20243dgs,jiang2023instantavatar}, for the evaluation on NeuMan dataset, we fit SMPL-X parameters of testing frames while freezing all other parameters with the image loss of Sec.~\ref{subsec:loss_functions}.

\begin{table}[t]
\footnotesize
\centering
\setlength\tabcolsep{1.0pt}
\def\arraystretch{1.1}
\caption{
Comparisons of 3D human avatars on the test set of X-Humans~\cite{shen2023x}.
Methods with * use additional depth maps for the training.
}
\label{table:compare_sota_xhumans}
\vspace*{-3mm}
\scalebox{0.8}{
\begin{tabular}{C{3.0cm}|C{1.2cm}C{1.2cm}C{1.2cm}|C{1.2cm}C{1.2cm}C{1.2cm}|C{1.2cm}C{1.2cm}C{1.2cm}}
\specialrule{.1em}{.05em}{.05em}
\multirow{2}{*}{Methods} & \multicolumn{3}{c|}{\textit{\textbf{00028}}} & \multicolumn{3}{c|}{\textit{\textbf{00034}}} & \multicolumn{3}{c}{\textit{\textbf{00087}}} \\
 & PSNR\textuparrow & SSIM\textuparrow & LPIPS\textdownarrow & PSNR\textuparrow & SSIM\textuparrow & LPIPS\textdownarrow & PSNR\textuparrow & SSIM\textuparrow & LPIPS\textdownarrow\\ \hline
X-Avatar~\cite{shen2023x}* & 28.57 & 0.976 & 0.026 & 28.05 & 0.965 & 0.035 & 30.89  & 0.970 & 0.030 \\
\textbf{ExAvatar (Ours)} & \textbf{30.58} & \textbf{0.981} & \textbf{0.018} & \textbf{28.75} & \textbf{0.966} & \textbf{0.029} & \textbf{32.01} & \textbf{0.972} & \textbf{0.025} \\
\specialrule{.1em}{.05em}{.05em}
\end{tabular}
}
\end{table}

\begin{figure}[t]
\begin{center}
\includegraphics[width=0.84\linewidth]{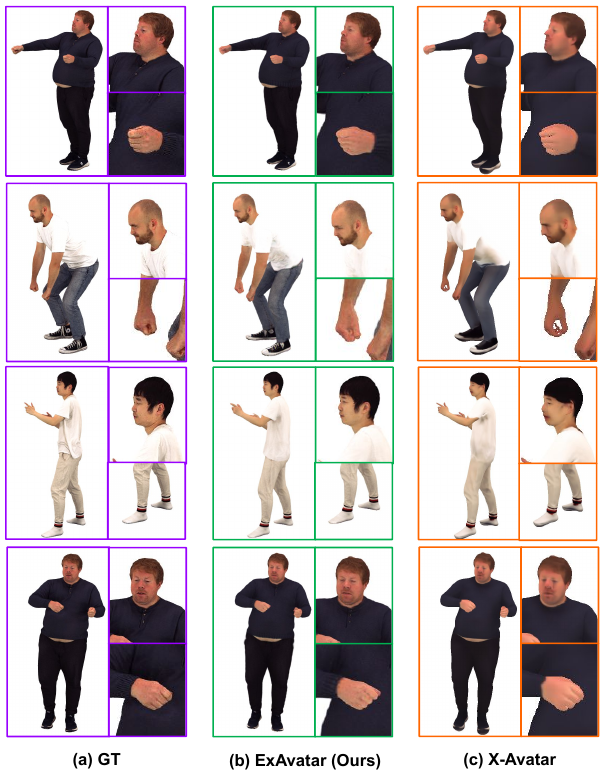}
\end{center}
\vspace*{-7mm}
\caption{
Qualitative comparison between our ExAvatar and X-Avatar~\cite{shen2023x} on the test set of X-Humans~\cite{shen2023x}.
}
\vspace*{-5mm}
\label{fig:compare_sota_xhumans}
\end{figure}

\begin{table}[t]
\setlength{\tabcolsep}{1pt}
\begin{minipage}{.45\linewidth}
\centering
\caption{Ablation study for the effectiveness of incorporating Laplacian regularizer to our 3D Gaussian-based system on the test set of NeuMan~\cite{jiang2022neuman}.}
\label{table:ablation_lap_reg}
\vspace*{-3mm}
\scalebox{0.7}{
\begin{tabular}{C{3.6cm}|C{1.2cm}C{1.2cm}C{1.2cm}}
\specialrule{.1em}{.05em}{.05em}
Settings & PSNR\textuparrow & SSIM\textuparrow & LPIPS\textdownarrow \\ \hline
Without Lap. reg. & 28.21 & 0.968 & 0.199 \\
\textbf{With Lap. reg. (Ours)} & \textbf{34.80} & \textbf{0.984} & \textbf{0.009} \\
\specialrule{.1em}{.05em}{.05em}
\end{tabular}
}
\end{minipage}\hfill
\begin{minipage}{.50\linewidth}
\centering
\caption{Ablation study for the effectiveness of our face loss on the cropped face images of a test set of X-Humans~\cite{shen2023x}.}
\label{table:ablation_face_loss}
\vspace*{-3mm}
\scalebox{0.7}{
\begin{tabular}{C{3.6cm}|C{1.2cm}C{1.2cm}C{1.2cm}}
\specialrule{.1em}{.05em}{.05em}
Settings & PSNR\textuparrow & SSIM\textuparrow & LPIPS\textdownarrow \\ \hline
Without face loss & 20.02 & 0.671 & \textbf{0.06} \\
\textbf{With face loss (Ours)} & \textbf{22.07} & \textbf{0.693} & \textbf{0.06}  \\
\specialrule{.1em}{.05em}{.05em}
\end{tabular}
}
\end{minipage}
\end{table}

Fig.~\ref{fig:compare_sota_neuman} shows that ours produces photorealistic renderings in novel views and poses.
For example, prints on the shirts (the first and third rows) are significantly sharper and clearer than those of previous works.
Most importantly, ours produces faces and hands in novel views and poses substantially better than previous avatars.
As previous avatars do not have controllability on faces and hands, the averaged blurry textures are baked in (faces in the first row and hands in the second row and fourth row).
On the other hand, ours has sharp textures benefiting from the whole-body modeling.

Tab.~\ref{table:compare_sota_xhumans} and Fig.~\ref{fig:compare_sota_xhumans} show that our ExAvatar outperforms previous whole-body avatar~\cite{shen2023x} on X-Humans~\cite{shen2023x} dataset even without using depth maps, while the previous work relies on it.
Our hybrid representation of the surface mesh and 3D Gaussians leads to stable training and shaper textures of faces and hands.
For X-Avatar's results, we used their officially released pre-trained weights and code.
Following Shen~\etal~\cite{shen2023x}, we used given SMPL-X parameters without further fitting them to testing frames.

\subsection{Ablation study}
In this section, we ablate the effectiveness of our hybrid representation of the surface mesh and 3D Gaussians, which enables us to incorporate Laplacian regularizer and face loss into our system.
Fig.~\ref{fig:lap_reg_effect} and Tab.~\ref{table:ablation_lap_reg} show that incorporating the Laplacian regularizer into our system brings significant performance boost and stability.
Fig.~\ref{fig:face_loss_effect} and Tab.~\ref{table:ablation_face_loss} show the benefit of the proposed face loss.
The numbers in Tab.~\ref{table:ablation_face_loss} are measured only for cropped face images when the face is visible to evaluate the effectiveness of the face loss.
The face visibility is decided by rasterizing SMPL-X meshes and checking the number of rasterized triangle faces of the face region.

\section{Conclusion}

\noindent\textbf{Summary.}
We present \textbf{ExAvatar}, an expressive whole-body 3D avatar that can be made from a short monocular video.
We propose a hybrid representation of the surface mesh and 3D Gaussians to address 1) the limited diversity of facial expressions and poses in the video and 2) the absence of 3D observations, such as 3D scans and RGBD images.
Our hybrid representation makes ExAvatar fully compatible with the facial expression space of SMPL-X and significantly reduces artifacts in novel facial expressions and novel poses.

\noindent\textbf{Limitations.}
First, as the inside of the mouth including the cavity and palm of the hands are often not observed in the video, our model hallucinates plausible geometry and textures.
Second, like previous avatars~\cite{peng2021neural,peng2021animatable,kwon2021neural,choi2022mononhr,jiang2022neuman,guo2023vid2avatar,jiang2023instantavatar,kocabas2023hugs,hu2023gaussianavatar,shen2023x,liu2024gea}, ours struggles in modeling dynamic clothes.
Material of clothes with motion information, such as velocity and acceleration, should be considered to properly model such dynamic clothes, out of our scope.

\noindent\textbf{Future works.}
To hallucinate unobserved human parts better, such as inside of the mouth, score distillation sampling~\cite{poole2023dreamfusion} can be used to \emph{generate} images and use them for supervision.
In addition, adding relightability to our ExAvatar is a promising and interesting future direction.

\newpage

\begin{center}
\textbf{\large Supplementary Material \textit{for} \\ \vspace{2mm}
\large{``Expressive Whole-Body 3D Gaussian Avatar"}}
\end{center}

\setcounter{section}{0}
\setcounter{table}{0}
\setcounter{figure}{0}

\renewcommand{\thesection}{\Alph{section}}   
\renewcommand{\thetable}{\Alph{table}}   
\renewcommand{\thefigure}{\Alph{figure}}

In this supplementary material, we provide more experiments, discussions, and other details that could not be included in the main text due to the lack of pages.
The contents are summarized below:
\begin{compactitem}
    \item Sec.~\ref{sec:co_registration_smplx}: Details of co-registration of SMPL-X
    \item Sec.~\ref{sec:mlp_architecture}: Detailed architecture of MLPs
    \item Sec.~\ref{sec:regularizers}: Regularizers
    \item Sec.~\ref{sec:running_time_compare}: Running time comparison
    \item Sec.~\ref{sec:geometry_comparison}: 3D geometry comparison
    \item Sec.~\ref{sec:compare_to_gen_ai}: Comparison to generative AIs
    \item Sec.~\ref{sec:implementation_details}: Implementation details
    \item Sec.~\ref{sec:failure_cases}: Failure cases
\end{compactitem}

\section{Details of co-registration of SMPL-X}~\label{sec:co_registration_smplx}
This section describes details of the co-registration of SMPL-X of Sec.~\textcolor{red}{3.1} of the main manuscript.

\noindent\textbf{FLAME registration.}
Given DECA~\cite{feng2021learning}'s output FLAME parameters (\emph{i.e.}, face shape parameter, facial expression code, and jaw pose), we further fit them to 2D keypoints of face, where the 2D keypoints are from an off-the-shelf regressor~\cite{mmpose2020}.
For the fitting, we minimize the below loss functions.
\begin{equation}
L_\text{FLAME} = L_\text{kpt} + 0.1 L_\text{init},
\end{equation}
where $L_\text{kpt}$ and $L_\text{init}$ represent 1) $L1$ distance between projected and target 2D keypoints and 2) $L1$ distance between optimizing FLAME parameters and the initial regressed ones, respectively.
In this way, we can make FLAME's meshes pixel-aligned with the image while preventing them from being too far from DEAC's regressed ones.

\noindent\textbf{SMPLX registration.}
Given Hybrid-X~\cite{li2023hybrikx}'s output SMPL-X parameters, we further fit them to 2D keypoints of the whole body, where the 2D keypoints are from an off-the-shelf regressor~\cite{mmpose2020}.
We replace the facial expression code of Hybrid-X with that of DECA as DECA's output is more accurate.
As FLAME and SMPL-X share the same facial expression space, we share the facial expression code of FLAME and SMPL-X during the registration.
Before performing the LBS inside of the SMPL-X layer, we 1) add the joint offset $\Delta \mathbf{J}$ to the T-pose joint locations of SMPL-X and 2) add the face offset $\Delta \mathbf{V}_\text{face}$ to the face region of the T-pose template mesh.
The joint offset allows us to obtain a more personalized 3D skeleton and surface than only using the SMPL-X shape parameter.
Please note that the joint offset affects both skeleton and surface as the added joint locations are used for following forward kinematics and LBS.

We minimize the below loss functions.
\begin{equation}
L_\text{SMPLX} = L_\text{kpt} + 0.1 L_\text{init} + L_\text{face} + L_\text{reg},
\end{equation}
\begin{equation}
L_\text{face} = 10 L_\text{vertex} + 10000 L_\text{lap} + L_\text{edge},
\end{equation}
\begin{equation}
L_\text{reg} = 0.01 L_\text{shape} + 100 L_\text{jo} + L_\text{sym},
\end{equation}
where $L_\text{kpt}$ and $L_\text{init}$ represent 1) $L1$ distance between projected and target 2D keypoints and 2) $L1$ distance between optimizing SMPL-X parameters and the initial regressed ones, respectively.
We compute the $L_\text{kpt}$ with and without the face offset $\Delta \mathbf{V}_\text{face}$ to prevent the face offset from representing any pose-dependent and expression-dependent things.
$L_\text{face}$ is to optimize the face offset $\Delta \mathbf{V}_\text{face}$.
$L_\text{vertex}$, $L_\text{lap}$, and $L_\text{edge}$ represent 1) $L1$ distance between SMPL-X face vertices and FLAME vertices, 2) $L2$ distance between Laplacian of SMPL-X face mesh and FLAME mesh, and 3) $L1$ distance between edge length of SMPL-X face mesh and FLAME mesh, respectively.
We compute $L_\text{vertex}$ with and without the pose and facial expression to prevent the face offset from representing any pose-dependent and expression-dependent things.
$L_\text{lap}$ and $L_\text{edge}$ are computed only for meshes without pose and facial expressions.
We compute $L_\text{face}$ only when the face is visible.
To check the face visibility, we compute two vectors.
First, a vector from the face center to the middle of the eyes in the $xz$ space of the camera-centered coordinate system.
Second, a vector from the camera position (origin) to the face center in $xz$ space of the camera-centered coordinate system.
We decide the face is visible if the dot product between the two vectors is smaller than $\text{cos}(135\text{°})$.
Finally, $L_\text{reg}$ is to regularize the SMPL-X shape parameter, joint offset, and face offset.
$L_\text{shape}$ is a squared $L2$ norm of the SMPL-X shape parameter.
$L_\text{jo}$ is a squared $L2$ norm of the joint offset, which prevents an extreme joint offset, and $L_\text{sym}$ is for symmetricity of the joint offset and face offset, similar to Feng~\etal~\cite{feng2021learning}.

\section{Architecture of MLPs}~\label{sec:mlp_architecture}
This section describes the detailed architecture of MLPs of Sec.~\textcolor{red}{3.2} of the main manuscript.
The MLPs are briefly described in Fig.~\textcolor{red}{4} of the main manuscript.

\subsection{MLPs without pose conditioning}~\label{subsec:mlp_wo_pose}
The MLPs without the pose conditioning (the red MLPs of Fig.~\textcolor{red}{4} of the main manuscript) consist of two types of MLPs.
The first MLP takes the triplane feature $\mathbf{F}$ and outputs the geometry of 3D Gaussian points (\emph{i.e.}, $\Delta \mathbf{V}_\text{tri}$ and $\mathbf{S}_\text{tri}$).
It consists of four fully connected layers with a hidden size of 128.
We use the group normalization~\cite{wu2018group} and ReLU activation function between each fully connected layer.
The second MLP takes the triplane feature $\mathbf{F}$ and outputs RGB colors of 3D Gaussian points $\mathbf{C}_\text{tri}$.
It has the same architecture as that of the first MLP except for the output dimension.

\subsection{MLPs with pose conditioning}
The MLPs with the pose conditioning (the blue MLPs of Fig.~\textcolor{red}{4} of the main manuscript) consist of two types of MLPs.
The MLPs have the same architecture as that of the MLPs of Sec.~\ref{subsec:mlp_wo_pose} except for the input and output dimensions.
The first MLP takes the triplane feature $\mathbf{F}$ and 3D pose $\theta$ of SMPL-X without the root pose and outputs geometry offsets of 3D Gaussian points (\emph{i.e.}, $\Delta \mathbf{V}_\text{pose}$ and $\Delta \mathbf{S}_\text{pose}$).
The second MLP takes the 1) triplane feature $\mathbf{F}$, 2) 3D pose $\theta$ of SMPL-X without the root, and 3) normal of each 3D Gaussian point and outputs RGB offset $\Delta \mathbf{C}_\text{pose}$.

\section{Regularizers}~\label{sec:regularizers}
This section describes regularizers, not introduced in Sec.~\textcolor{red}{3.4} of the main manuscript.
First, to prevent 3D Gaussian points from being too far from underlying canonical mesh $\bar{\mathbf{V}}$, we penalize squared $L2$ norm of 3D Gaussian offsets $\Delta \mathbf{V}_\text{tri}$ and $\Delta \mathbf{V}_\text{pose}$.
Second, to prevent 3D Gaussian from being too big, we penalize the squared $L2$ norm of 3D Gaussian scales.
Third, as hands often suffer from noisy colors due to their small scales and complicated articulations, we minimize the squared $L2$ distance between RGB colors of hand 3D Gaussian points and the average colors of hand 3D Gaussian points.
Finally, we use the same $L_\text{jo}$ of Sec.~\ref{sec:co_registration_smplx}.

\begin{table}[t]
\footnotesize
\centering
\setlength\tabcolsep{1.0pt}
\def\arraystretch{1.1}
\caption{
Frames per second comparisons of 3D human avatars.
}
\label{table:running_time_compare}
\vspace*{-3mm}
\scalebox{0.8}{
\begin{tabular}{C{2.0cm}C{2.3cm}C{2.0cm}C{3.0cm}C{6.0cm}}
\specialrule{.1em}{.05em}{.05em}
NeuMan~\cite{jiang2022neuman} & Vid2Avatar~\cite{guo2023vid2avatar} & X-Avatar~\cite{shen2023x} & \textbf{ExAvatar (Ours)} & \textbf{ExAvatar (Ours wo. pose cond.)} \\ \hline
 0.02 & 0.04 & 0.12 & 26 & 47 \\
\specialrule{.1em}{.05em}{.05em}
\end{tabular}
}
\end{table}

\section{Running time comparison}~\label{sec:running_time_compare}
Tab.~\ref{table:running_time_compare} shows that ours achieves real-time animation and rendering speed while existing works~\cite{jiang2022neuman,guo2023vid2avatar} fail to.
Without the pose conditioning (the blue MLP of Fig.~\textcolor{red}{4} of the main manuscript), ours achieves even faster speed.
For all methods, we measure the running time only for the human animation and rendering without the background rendering.
A single NVIDIA V100 GPU is used, and the rendering resolution is 1024 $\times$ 1024.

\begin{figure}[t]
\begin{center}
\includegraphics[width=0.7\linewidth]{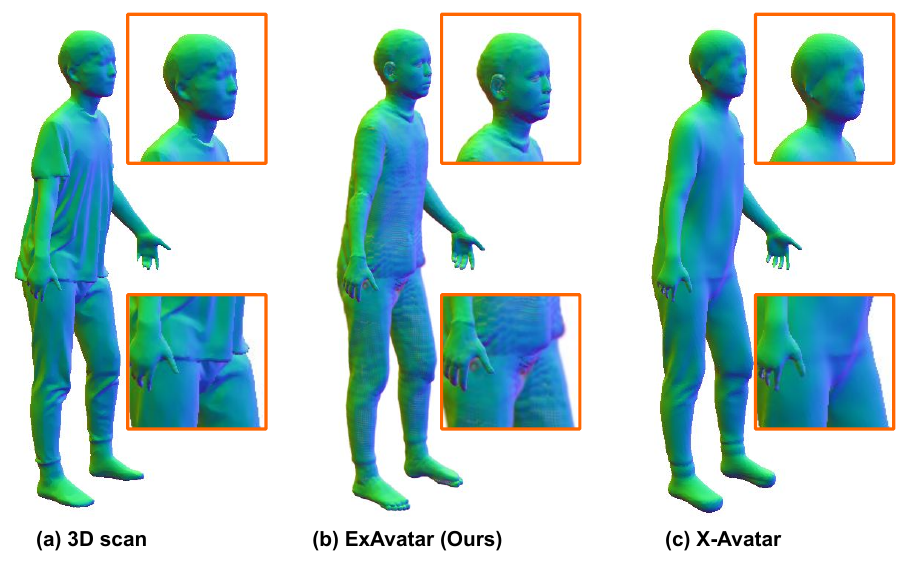}
\end{center}
\vspace*{-7mm}
\caption{
Qualitative comparison of rendered normals from (a) our ExAvatar and (b) X-Avatar~\cite{shen2023x} on the test set of X-Humans~\cite{shen2023x}.
}
\label{fig:compare_geometry}
\end{figure}

\begin{table}[t]
\footnotesize
\centering
\setlength\tabcolsep{1.0pt}
\def\arraystretch{1.1}
\caption{
3D geometry comparison between our ExAvatar and X-Avatar~\cite{shen2023x} on the subset of the test set of X-Humans~\cite{shen2023x}.
}
\label{table:geometry_compare}
\scalebox{0.8}{
\begin{tabular}{C{3.5cm}|C{2.5cm}C{3.5cm}C{4.0cm}}
\specialrule{.1em}{.05em}{.05em}
Methods & Mask IoU \textuparrow & Depthmap (mm) \textdownarrow & Normal consistency \textuparrow \\ \hline
X-Avatar & 94.13 & 11.24 & 0.823 \\
\textbf{ExAvatar (Ours)} & \textbf{96.11} & \textbf{10.97} & \textbf{0.868} \\
\specialrule{.1em}{.05em}{.05em}
\end{tabular}
}
\end{table}

\section{Geometry comparison}~\label{sec:geometry_comparison}
Fig.~\ref{fig:compare_geometry} and Tab.~\ref{table:geometry_compare} show that our ExAvatar produces better 3D geometry than previous state-of-the-art 3D whole-body avatar~\cite{shen2023x}, especially around the face and neck.
It is not straightforward to extract actual 3D geometry from 3D Gaussians.
Instead, we render masks/depth maps/normal maps of Gaussians to multiple viewpoints and comparing them with ground truth.
However, when rendering depth/normals, even recent SOTA methods~\cite{guedon2024sugar} simply take the depth/normal from the center of Gaussians and alpha-blend it in the screen space.
This may result in incorrect depth/normal maps as elongated Gaussians can deviate from the depth/normal values of the Gaussian centers.
Nevertheless, we report 3D geometry comparisons in the figure and table for reference.
We leave actual 3D geometry extraction from ExAvatar as a future work.

\begin{figure}[t]
\begin{center}
\includegraphics[width=0.7\linewidth]{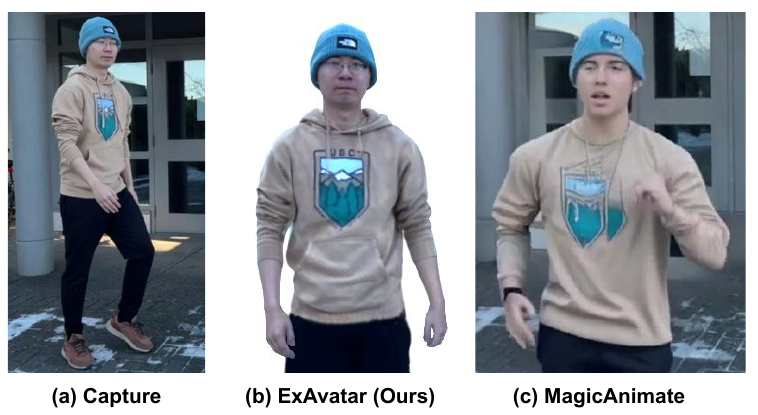}
\end{center}
\vspace*{-7mm}
\caption{
Qualitative comparison of (b) our ExAvatar and (c) state-of-the-art generative method~\cite{xu2023magicanimate} for the human animation.
}
\vspace*{-5mm}
\label{fig:compare_gen_ai}
\end{figure}

\section{Comparison to generative AIs}~\label{sec:compare_to_gen_ai}
Motivated by the powerful modeling capability of the diffusion models~\cite{rombach2022high,ho2020denoising}, several human animation methods~\cite{xu2023magicanimate} have been introduced.
They can animate humans only from a single image using DensePose~\cite{guler2018densepose} or 2D pose as the driving signals.
Despite their photorealistic and plausible outputs, they lack authenticity, as shown in Fig.~\ref{fig:compare_gen_ai}.
The person of the (b) generated image from MagicAnimate~\cite{xu2023magicanimate} has definitely a different face identity and clothes compared to the (a) captured image, although they look similar.
The result of MagicAnimate~\cite{xu2023magicanimate} is obtained using their official released code.
This is because of the hallucination, one of the biggest challenges for modeling generative AIs.
On the other hand, ours might lack plausibility, especially for unobserved human parts such as inside of mouth and cloth dynamics; however, ours has more authenticity.
We think combining the plausibility of generative approaches and the authenticity of our approach should be an interesting and future direction, as discussed in Sec.~\textcolor{red}{5} of the main manuscript.
\section{Implementation details}~\label{sec:implementation_details}
PyTorch~\cite{paszke2017automatic} is used for implementation. 
For the training, we use Adam optimizer~\cite{kingma2014adam}.
The initial learning rate is set to $10^{-3}$ and reduced by a factor of 10 at the 75 \% and 90 \% of total number of iterations.
We use a single V100 NVIDIA GPU for experiments, and the training takes 1.5 hours to 5 hours depending on the duration of the videos.
All other details will be available in our code. 
\begin{figure}[t]
\begin{center}
\includegraphics[width=0.7\linewidth]{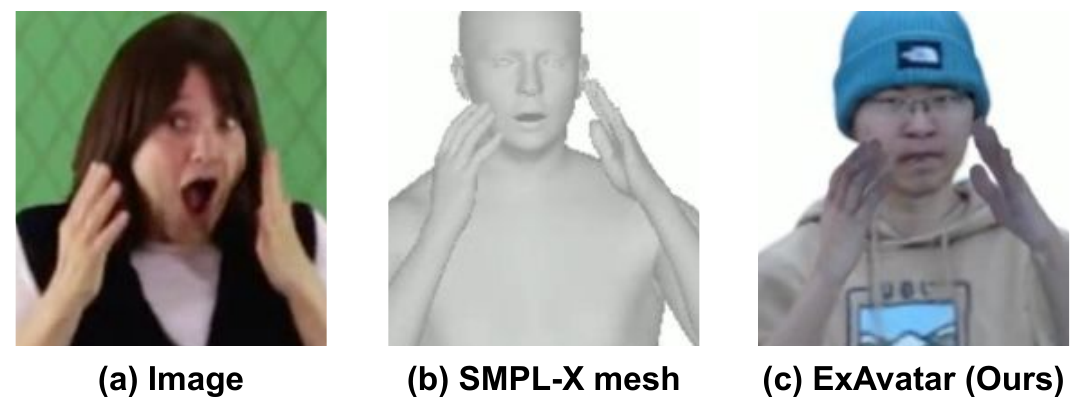}
\end{center}
\vspace*{-7mm}
\caption{
Failure case of our ExAvatar due to (b) the wrong regressed facial expression code $\psi$ from the off-the-shelf regressor~\cite{feng2021learning}.
}
\vspace*{-5mm}
\label{fig:failure_case_wrong_param}
\end{figure}

\begin{figure}[t]
\begin{center}
\includegraphics[width=0.7\linewidth]{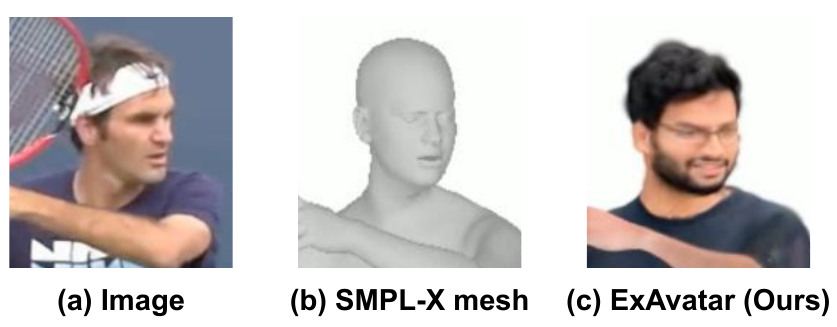}
\end{center}
\vspace*{-7mm}
\caption{
Failure case of our ExAvatar due to the baked facial appearance of a subject.
}
\vspace*{-5mm}
\label{fig:failure_case_baked_expr}
\end{figure}

\section{Failure cases}~\label{sec:failure_cases}
Fig.~\ref{fig:failure_case_wrong_param} shows that our rendered avatar does not have a smiling facial expression despite the person in the image is smiling.
This is because of the wrong regressed facial expression code $\psi$ from the off-the-shelf regressor~\cite{feng2021learning} as (b) rendered SMPL-X mesh also has the wrong facial expression.
As our ExAvatar is animated with the regressed facial expression code, ours also has wrong facial expression.
We think advanced facial models and high-fidelity regressors should be investigated to address this failure case.

Fig.~\ref{fig:failure_case_baked_expr} additionally shows our failure case.
Although the person in (a) the image is frowning, our rendered avatar is slightly smiling.
This is because the subject of our avatar in the NeuMan dataset~\cite{jiang2022neuman} is always smiling during the short video capture.
Hence, the smiling facial appearance is baked into our avatar.
Canonicalizing facial appearance from a short monocular is greatly challenging.
In particular, when the face takes a few pixels in the video like our case, 2D keypoints are often noisy, which makes the lip geometry of SMPL-X/FLAME are not fit perfectly in the co-registration stage (Sec.~\textcolor{red}{3.1} of the main manuscript).
Such a misalignment error is propagated to our ExAvatar learning stage, which causes the failure case.
We think using priors of the canonical facial appearances using generative methods can be one way to address this failure case.

\newpage

\bibliographystyle{splncs04}
\bibliography{main}
\end{document}